\documentclass{article}

\usepackage[final]{colm2026_conference}

\usepackage[utf8]{inputenc} %
\usepackage[T1]{fontenc}    %
\usepackage[hidelinks]{hyperref}       %
\usepackage{url}            %
\usepackage{booktabs}       %
\usepackage{amsfonts}       %
\usepackage{amsmath}
\usepackage{nicefrac}       %
\usepackage{microtype}      %
\usepackage{xcolor}         %
\usepackage{fontawesome5} %

\usepackage{graphicx}
\usepackage{multirow}
\usepackage{xspace}
\usepackage[most]{tcolorbox}
\usepackage{mdframed}
\usepackage{colortbl}
\usepackage{array}
\usepackage[normalem]{ulem}
\usepackage{placeins}

\usepackage{amsmath,amsfonts,bm}

\def\eqref#1{equation~\ref{#1}}

\def\1{\bm{1}}

\DeclareMathAlphabet{\mathsfit}{\encodingdefault}{\sfdefault}{m}{sl}
\SetMathAlphabet{\mathsfit}{bold}{\encodingdefault}{\sfdefault}{bx}{n}

\definecolor{bestcolor}{HTML}{E6E6FA}
\definecolor{secondcolor}{HTML}{E5F3FA}
\definecolor{proprietary_model_color}{HTML}{FFFAFC}
\definecolor{open_source_model_color}{HTML}{FFFFFA}
\definecolor{skills_model_color}{HTML}{F8F8FF}
\newcommand{\best}[1]{\cellcolor{bestcolor}\textbf{#1}}
\newcommand{\second}[1]{\cellcolor{secondcolor}\uline{#1}}
\definecolor{appleTeal}{HTML}{00D2E0}
\definecolor{applePink}{HTML}{FF375F}
\definecolor{appleGray}{HTML}{8E8E93}

\newcommand{\model}{\mbox{\textsc{RewardHarness}}\xspace}

\title{RewardHarness: Self-Evolving Agentic Post-Training}

\author{%
  \textbf{Yuxuan Zhang$^{1,2,3,5,*}$, Cong Wei$^{2,4,*}$, Penghui Du$^{3,*}$, Bo Li$^{3,*}$, Junwen Miao$^{8}$}, \\[3pt]
  \textbf{Huaisong Zhang$^{6}$, Songcheng Cai$^{4}$, Yubo Wang$^{2,4}$, Dongfu Jiang$^{2,4,\dagger}$}, \\[3pt]
  \textbf{Yuyu Zhang$^{7}$, Changqian Yu$^{3,\S}$, Ping Nie$^{4,\dagger}$, Wenhu Chen$^{2,4,\dagger}$, Kelsey R Allen$^{1,2,\dagger}$}
  \\[10pt]
  \normalsize
  $^{1}$University of British Columbia \quad $^{2}$Vector Institute \quad $^{3}$KlingAI Research \\[2pt]
  $^{4}$University of Waterloo \quad $^{5}$Etude AI \quad $^{6}$Tsinghua University \\[2pt]
  $^{7}$Georgia Institute of Technology \quad $^{8}$Carnegie Mellon University
}

\begin{document}

\maketitle
\lhead{Published as a conference paper at COLM 2026}
{\renewcommand{\thefootnote}{\fnsymbol{footnote}}%
\footnotetext[1]{Equal Contribution. \quad $^{\S}$\,Project Lead. \quad $^{\dagger}$\,Advisors.}}

\begin{abstract}
Aligning generative models with human preferences remains a central challenge. While RLHF is widely adopted, it relies on large-scale preference data. This limitation is particularly evident in evaluating instruction-guided image edits, even though humans can often learn such preferences from only a few examples.
We present \textbf{\model}, a \textbf{ training-free, self-evolving} agentic reward framework that reframes reward modeling as context evolution rather than weight optimization. 
Instead of learning from large-scale annotations, \model aligns with human preferences by iteratively evolving a \textbf{library of tools and skills} from as few as 100 preference demonstrations.
Given a source image, an edited image, and an editing instruction, an orchestrator agent selects the most relevant subset of tools and skills from the maintained library, and sub-agents invoke them to construct a reasoning chain that produces a reward score. By comparing predicted scores with ground-truth preferences and analyzing successes and failures in the reasoning process, the orchestrator automatically refines its library of tools and skills without additional human annotation.
Using only 0.05\% of the EditReward preference data, \model achieves 47.4\% average accuracy on EditReward-Bench $K$=2/3/4 and GenAI-Bench, surpassing GPT-5 by 5.3 points. When used as a reward signal in FlowGRPO, RL-tuned models achieve 3.52 on ImgEdit-Bench. 
\textbf{Code and model are available at}
\faGlobe\ \textbf{Project Page:} 
\href{https://rewardharness.com/}{\textcolor{blue}{https://rewardharness.com/}}
\end{abstract}

\begin{figure}[t]
\centering
\vspace{-0.4em}
\includegraphics[width=0.97\textwidth]{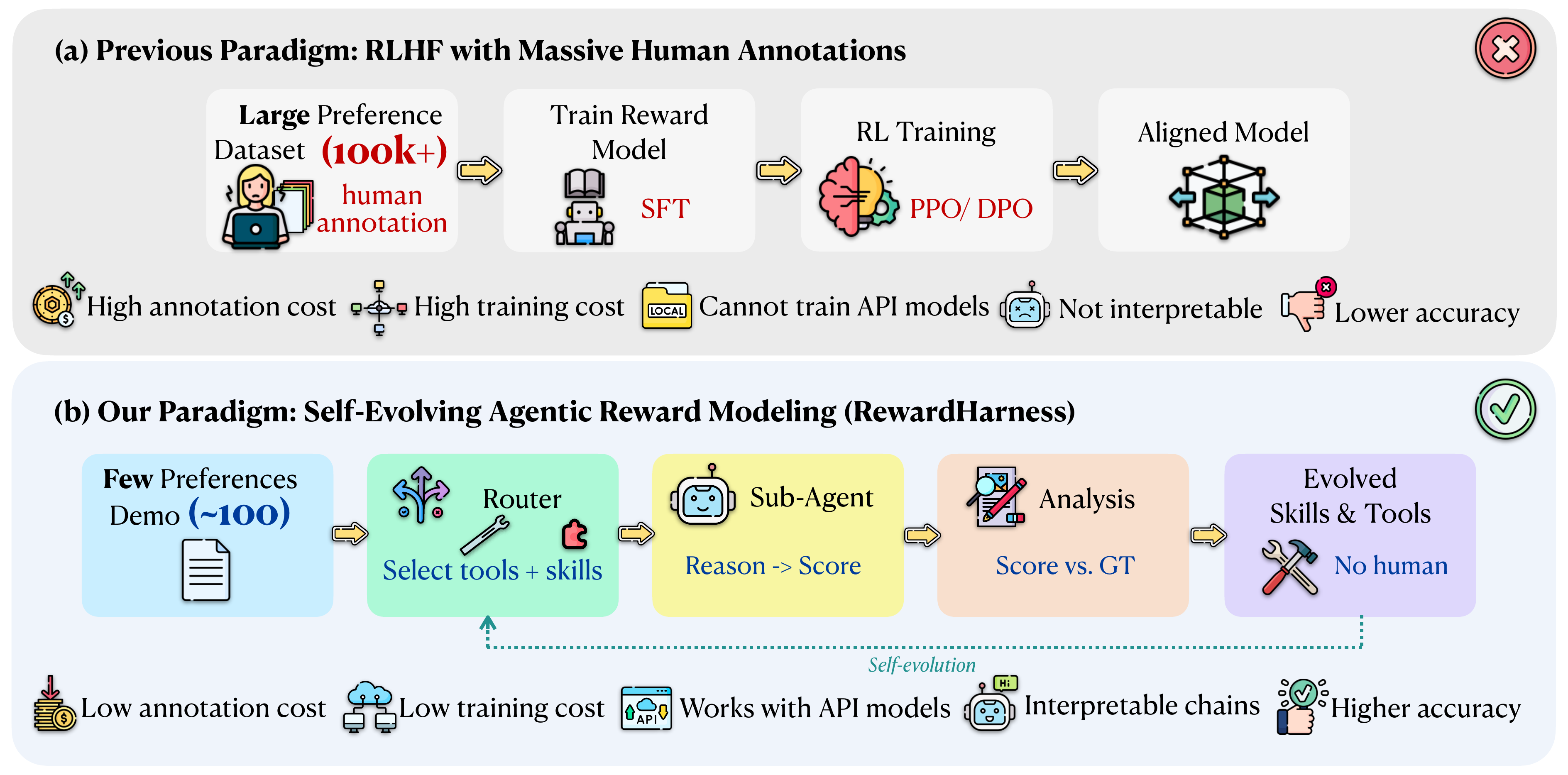}
\vspace{-0.8em}
\caption{\textbf{Paradigm comparison.} The conventional paradigm collects large-scale human preference annotations, typically around 100K examples, to train an evaluator model via supervised fine-tuning, whereas our training-free method learns human evaluation preferences from only 100 demonstrations.}
\label{fig:paradigm_comparison}
\end{figure}

\section{Introduction}
\label{sec:introduction}

Image editing has advanced rapidly, but \textbf{reliable evaluation remains a central bottleneck}. This challenge is even more pronounced in reinforcement learning for visual generation and editing, where progress depends on reward signals that faithfully reflect human preferences~\citep{flowgrpo,editr1,xue2025dancegrpo,chen2025blip3o,diffusionnft}.

As illustrated in Figure~\ref{fig:paradigm_comparison}(a), existing approaches~\citep{imagereward,kirstain2023pick,unifiedreward,xu2024visionreward,gong2025onereward,editreward,luo2025editscore,he2024videoscore,liu2025improving,richhf,wang2025worldpm} largely address this problem by collecting large-scale human preference annotations and training dedicated reward models on top of them. While effective, this paradigm is expensive and inflexible: it incurs substantial annotation cost, requires additional model training, often produces opaque scalar rewards, and is difficult to apply to closed or API-only foundation models. These limitations are particularly severe for image editing, where preference judgments are subtle, multi-dimensional, and depend on jointly understanding the editing instruction, the source image, and the edited result.

More importantly, it reveals a striking asymmetry. Human annotators can often internalize the target evaluation criteria from only a small calibration set and then apply them consistently at scale, whereas current models typically require hundreds of thousands of labeled comparisons to acquire similar preference behavior.
This raises the central question of this paper: \textbf{if humans can acquire image-editing preferences from a handful of demonstrations, can models do the same purely in context and without any parameter updates?}

We answer this question with \model, a self-evolving agentic reward framework that reframes reward modeling as context evolution rather than weight optimization. 
As illustrated in Figure~\ref{fig:paradigm_comparison}(b), the key idea is not to spend a small number of demonstrations on training a smaller reward model, but to use them to iteratively build an explicit and reusable library of evaluation knowledge. 
Specifically, \model evolves a library of \emph{Skills} and \emph{Tools}:
\emph{Skills} provide structured evaluation guidelines that break image-editing quality into fine-grained criteria, while \emph{Tools} provide structured specifications for targeted visual analysis, describing what should be checked, how it should be analyzed, and when the procedure should be invoked.
Given a source image, candidate edits, and an editing instruction, an orchestrator retrieves the most relevant subset of Skills and Tools, and a sub-agent composes them into an interpretable reasoning chain that produces a reward judgment.

This design leads to a different way of obtaining reward capability. Instead of fitting a monolithic reward network from massive annotations, \model uses only about 100 preference demonstrations to iteratively evaluate predictions against human labels, analyze successes and failures, and refine the underlying library without additional human supervision. In this sense, \textbf{\model is not merely a better reward model; it is a different way to obtain reward capability}. The resulting reward system is data-efficient, compatible with frozen and API-based models, and more interpretable because its evaluation behavior is externalized into editable Skills, Tools, and reasoning traces rather than hidden in model parameters.

\textbf{Key results.} Built on top of off-the-shelf foundation models, \model achieves strong performance without gradient-based reward-model training. With a Claude-based Orchestrator and a frozen Qwen2.5-VL-7B sub-agent, \model surpasses existing Qwen-based reward models trained with supervised fine-tuning on 200K preference pairs while using only 0.05\% of the preference data. \model (Gemini) achieves 47.4\% average accuracy on EditReward-Bench and GenAI-Bench, surpassing GPT-5 by 5.3 points. When used as a reward signal in FlowGRPO, \model-tuned models achieve 3.52 on ImgEdit-Bench.

\section{Method}
\label{sec:method}

\begin{figure*}[htbp]
\centering
\includegraphics[width=\textwidth]{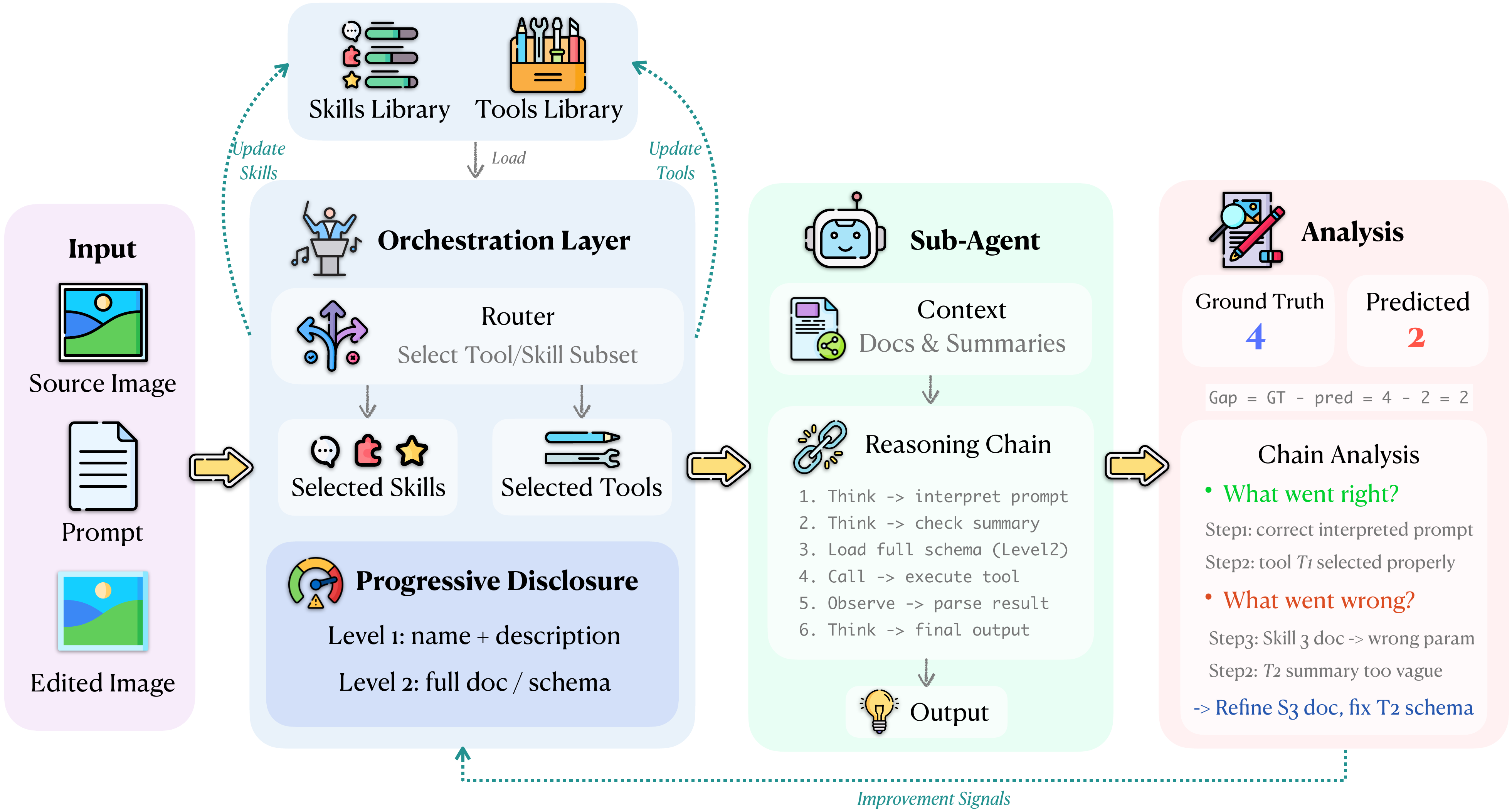}
\caption{\textbf{Overview of the \model self-evolution pipeline.} \textbf{(a)}~Multi-modal inputs (source image, edited images, editing prompt) are fed into the Orchestrator. \textbf{(b)}~The Orchestrator selects relevant entries from the Skills and Tools libraries via progressive disclosure routing. \textbf{(c)}~The Sub-Agent (a frozen VLM, e.g., Qwen2.5-VL-7B) builds a reasoning chain using selected skills and tools, producing a preference judgment. \textbf{(d)}~Outputs are scored against ground truth; the Orchestrator analyzes reasoning chains to generate improvement signals that update the libraries.}
\label{fig:pipeline}
\end{figure*}

We present \model, a self-evolving agentic reward system that acquires human evaluation preferences through context evolution alone, without updating any model parameters.
\model consists of two main components: an \textbf{Orchestrator} agent and a shared \textbf{Library} of interpretable evaluation artifacts.
At inference time, the Orchestrator retrieves relevant artifacts from the Library and injects them into the context of a frozen Sub-Agent VLM, which performs the preference judgment.
At evolution time, the Orchestrator drives iterative Library refinement using a small calibration set of human preference demonstrations.
Figure~\ref{fig:pipeline} provides an overview of the full pipeline.
We describe each component in turn: the problem formulation~(\S\ref{subsec:formulation}), the Skills and Tools Library~(\S\ref{subsec:libraries}), the Orchestrator~(\S\ref{subsec:orchestrator}), the Sub-Agent~(\S\ref{subsec:subagent}), and the self-evolution loop~(\S\ref{subsec:evolution}).

\subsection{Problem Formulation}
\label{subsec:formulation}

Given a source image $I_s$, an editing instruction $p$, and $K$ candidate edited images $\{I_1, \ldots, I_K\}$, the task is to produce a preference ranking $\pi$ over $\{1,\ldots,K\}$ such that $I_{\pi(1)} \succ I_{\pi(2)} \succ \cdots \succ I_{\pi(K)}$.
In \model, ranking is realized by a frozen VLM $\mathcal{M}$ steered entirely by a context $\mathcal{C}$ assembled at inference time:
\begin{equation}
\pi = \mathcal{M}\bigl(I_s,\; \{I_k\}_{k=1}^{K},\; p,\; \mathcal{C}\bigr),
\label{eq:ranking}
\end{equation}
where $\mathcal{C}$ comprises the Skill documents and Tool specifications selected by the Orchestrator; the parameters of $\mathcal{M}$ are never updated.

\subsection{Skills and Tools Library}
\label{subsec:libraries}

\begin{figure*}[htbp]
\centering
\includegraphics[width=\textwidth]{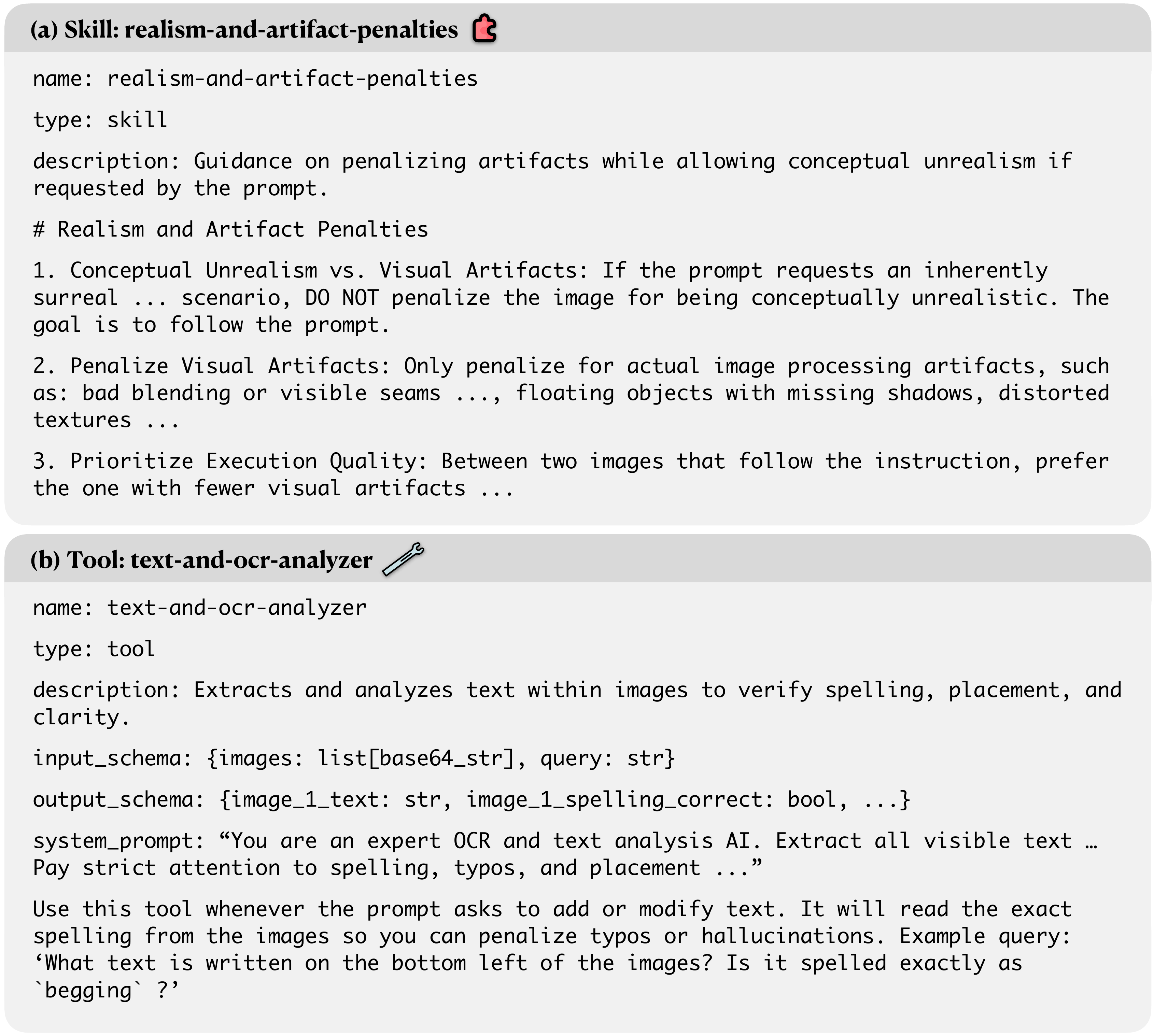}
\caption{Examples of a \textbf{Skill} and a \textbf{Tool} sampled from the Library at evolution iteration~69. Skills are declarative rubrics guiding the Sub-Agent's assessment criteria; Tools are procedural specifications instructing the Sub-Agent to perform targeted visual analysis.}
\label{fig:skill_tool_example}
\end{figure*}

\model maintains a \textbf{Library}, a versioned collection of Skills and Tools that encodes accumulated evaluation knowledge. The Library is initialized empty and grows through self-evolution (\S\ref{subsec:evolution}).
Representative examples of both components are shown in Figure~\ref{fig:skill_tool_example}.

\paragraph{Skills.}
A \emph{Skill} is a structured Markdown evaluation guideline containing: a \textbf{name}, a one-line \textbf{description}, a \textbf{scoring rubric} decomposing quality into assessable criteria, and \textbf{examples} illustrating correct application.
For instance, the skill \emph{realism-and-artifact-penalties} provides rubrics that distinguish visual artifacts (always penalized) from conceptual unrealism (acceptable when explicitly requested by the editing instruction).

\paragraph{Tools.}
A \emph{Tool} is a structured Markdown document that specifies a targeted visual analysis procedure: it defines the tool's \textbf{name}, \textbf{purpose}, expected \textbf{inputs} and \textbf{outputs}, \textbf{invocation conditions}, and a step-by-step \textbf{execution protocol}.
Unlike Skills (which provide declarative evaluation criteria), Tools provide \emph{procedural} specifications: by reading a Tool document, a general-purpose VLM can temporarily act as a specialized expert for a particular visual analysis task, without any parameter updates.
For example, the \emph{text-and-ocr-analyzer} tool instructs the Sub-Agent to extract, compare, and verify text content in source and edited images, catching typos and placement errors that holistic evaluation routinely misses.

\subsection{Orchestrator Layer}
\label{subsec:orchestrator}

The Orchestrator is a Claude-based LLM that serves two roles. During \emph{inference}, it examines the editing instruction and source image, then selects the appropriate Skills and Tools from the library to assemble the evaluation context $\mathcal{C}$ for the Sub-Agent. During \emph{evolution}, it analyzes the Sub-Agent's reasoning chains against ground-truth labels, performs root-cause analysis on errors, and proposes library updates (\S\ref{subsec:evolution}).

\subsection{Sub-Agent}
\label{subsec:subagent}

The Sub-Agent is a frozen, pluggable VLM that receives the multimodal inputs $I_s$, $\{I_k\}_{k=1}^{K}$, $p$, and the assembled context $\mathcal{C}$ from the Orchestrator.
By reading the Skill and Tool documents in $\mathcal{C}$, the Sub-Agent temporarily adopts the role of a specialized evaluator and constructs a structured reasoning chain.
Our default configuration uses Qwen2.5-VL-7B-Instruct, but the Sub-Agent is fully pluggable: we also evaluate Gemini as a drop-in replacement (Table~\ref{tab:editreward}).
The reasoning chain proceeds in three steps:

\begin{enumerate}
    \item \textbf{Rubric application.} For each Skill in $\mathcal{C}$, the Sub-Agent applies its scoring rubric to every candidate image, producing per-criterion assessments grounded in the skill's guidelines and examples.
    \item \textbf{Tool execution (optional).} For each Tool in $\mathcal{C}$ whose invocation conditions are met, the Sub-Agent follows the tool's execution protocol to perform a targeted visual analysis (e.g., OCR extraction, spatial relationship verification, object counting) and appends the structured result to the reasoning chain.
    \item \textbf{Aggregation and ranking.} The Sub-Agent synthesizes all per-criterion assessments and tool outputs into a final preference ranking $\pi$ over the $K$ candidates.
\end{enumerate}

\subsection{Self-Evolution Loop}
\label{subsec:evolution}

The self-evolution loop takes as input a small calibration set of $N=100$ human preference demonstrations $\mathcal{D} = \{(I_s^{(i)}, p^{(i)}, \{I_k^{(i)}\}, \pi^{*(i)})\}_{i=1}^{N}$.
The Orchestrator partitions $\mathcal{D}$ into a training split $\mathcal{D}_{\mathrm{train}}$ (${\approx}70$ examples) and a held-out validation split $\mathcal{D}_{\mathrm{val}}$ (${\approx}30$ examples).
Each iteration of the loop proceeds through five stages:

\paragraph{Step 1: Evaluation.}
For each example in $\mathcal{D}_{\mathrm{train}}$, the Orchestrator retrieves the most relevant Skills and Tools from the current Library and assigns them to a Sub-Agent.
The Sub-Agent constructs a reasoning chain and produces a predicted preference ranking $\hat{\pi}^{(i)}$ following the procedure in \S\ref{subsec:subagent}.

\paragraph{Step 2: Scoring.}
Predicted rankings are compared against ground-truth human preferences; samples are partitioned into correct predictions and errors.

\paragraph{Step 3: Chain analysis.}
The Orchestrator examines reasoning chains from both correct and incorrect predictions.
For errors, it performs root-cause analysis: identifying whether the failure stems from a missing evaluation criterion (suggesting a new Skill), an incorrect rubric application (suggesting a Skill modification), or a perceptual hallucination (suggesting a new or improved Tool).
For correct predictions, it identifies which Skills and Tools were instrumental, reinforcing their retention.
The Chain Analyzer produces a structured improvement proposal rather than unstructured advice.
Each proposal identifies a target artifact, specifies one of three operations: creating, modifying, or deprecating a Skill or Tool, and provides a justification grounded in the observed correct and incorrect reasoning traces.
The proposal does not directly modify the committed Library.
Instead, it is applied to a candidate snapshot that is subsequently evaluated by the validation gate.

\paragraph{Step 4: Library update.}
Based on the analysis, the Orchestrator proposes one of three actions: (i)~\emph{creating} a new Skill or Tool, (ii)~\emph{modifying} an existing entry, or (iii)~\emph{deprecating} an entry that consistently leads to incorrect reasoning.
In addition to incremental updates, the system can also perform aggressive pruning to remove accumulated artifacts from the exploration phase. In our experiments, the pruning phase begins around iteration~50 after the library peaks at 13 entries (8~Skills + 5~Tools), eventually producing a compact final library with 7 entries (3~Skills + 4~Tools).

\paragraph{Step 5: Validation and gating.}
The candidate Library is evaluated on the validation split $\mathcal{D}_{\mathrm{val}}$.
An update is committed only when its validation accuracy strictly exceeds that of the current committed Library.
Otherwise, the candidate is discarded and the current Library snapshot is restored.
Thus, proposed edits do not survive solely because they are recommended by the Orchestrator.
This conservative gating mechanism prevents regression.
We also observed markedly different acceptance rates for Skills and Tools: only 8.1\% of Skill proposals were accepted, reflecting the difficulty of modifying declarative rubrics without regression, compared to 41.4\% for Tool proposals, reflecting the modularity of procedural capabilities.
The loop terminates after a fixed budget of iterations.
The final committed Library is then frozen and used without further updates.
In our experiments, this final Library (3~Skills + 4~Tools) achieved 62.5\% validation accuracy, a 47\% relative improvement over the 42.5\% empty-library baseline.

\vspace{-.2cm}
\begin{table*}[t!]
\centering
\small
\renewcommand{\arraystretch}{1.3}
\setlength{\tabcolsep}{5.0pt}
\begin{tabular}{l | w{c}{3.2em} w{c}{3.2em} w{c}{3.2em} w{c}{5em} | w{c}{3.2em} w{c}{3.2em}}
\toprule
\multirow{2}{*}{\textbf{Method}} & \multicolumn{3}{c}{\textbf{EditReward-Bench}} & \multirow{2}{*}{\textbf{GenAI-Bench}} & \multirow{2}{*}{\textbf{Avg.}} & \multirow{2}{*}{$\boldsymbol{\Delta}$} \\
\cmidrule(lr){2-4} & $K$=2 & $K$=3 & $K$=4 &  &  &  \\
\midrule
\rowcolor{proprietary_model_color}
\multicolumn{7}{c}{\textit{Proprietary Models}} \\
\midrule
GPT-4o            & 45.7 & 27.3 & 7.3  & 53.5 & 33.5 & -- \\
GPT-5             & 57.5 & 38.5 & \second{12.8} & 59.6 & 42.1 & +8.6 \\
Gemini-2.0-Flash  & 52.4 & 33.3 & \best{13.5}  & 53.3 & 38.1 & +4.6 \\
Gemini-2.5-Flash  & \second{58.6} & 39.9 & 12.2 & 57.0 & 41.9 & +8.4 \\
Claude-Haiku-4.5  & 57.9 & 30.7 & 7.4  & 47.1 & 35.8 & +2.3 \\
\midrule
\rowcolor{open_source_model_color}
\multicolumn{7}{c}{\textit{Open-Source Models}} \\
\midrule
Qwen2.5-VL-7B     & 52.7 & 24.7 & 3.4  & 40.5 & 30.3 & $-$3.2 \\
Qwen2.5-VL-32B    & 50.5 & 25.3 & 4.1  & 39.3 & 29.8 & $-$3.7 \\
MiMo-VL-7B        & 49.5 & 30.4 & 9.5  & 57.9 & 36.8 & +3.3 \\
EditReward (Qwen2.5-VL-7B)  & 57.0 & 36.0 & 10.8 & 64.0 & 42.0 & +8.5 \\
EditReward (MiMo)  & 56.5 & 42.7 & 11.5 & \second{65.7} & 44.1 & +10.6 \\
\midrule
\rowcolor{skills_model_color}
\multicolumn{7}{c}{\textit{Models with Evolving Skills and Tools (Ours)}} \\
\midrule
Ours (Qwen2.5-VL-7B Subagent)    & 57.9 & \best{46.7} & 10.8 & \best{67.5} & \second{45.7} & +12.2 \\
Ours (Gemini-2.0-Flash Subagent)  & \best{66.2} & \second{45.3} & \best{13.5} & 64.4 & \best{47.4} & +13.9 \\
\bottomrule
\end{tabular}
\caption{\textbf{Comparison of image editing evaluators on editing reward benchmarks.} \colorbox{bestcolor}{\textbf{Best}} and \colorbox{secondcolor}{\uline{second-best}} results are highlighted. $\Delta$ measures the average improvement over the GPT-4o baseline. Despite using the same Qwen2.5-VL-7B backbone, \model surpasses EditReward, which is fine-tuned on 200K preference examples.}
\label{tab:editreward}
\end{table*}

\section{Experiments}
\label{sec:experiments}

We evaluate \model on editing reward benchmarks and downstream RL applications.
The default sub-agent is a frozen Qwen2.5-VL-7B-Instruct backbone served via vLLM; no model parameters are updated at any point.
We also evaluate Gemini as a drop-in sub-agent replacement (Table~\ref{tab:editreward}).

\subsection{Main Results on Image-Editing Evaluation}
\label{subsec:preference_modeling}

We evaluate preference judgment accuracy on two established benchmarks: EditReward-Bench~\citep{editreward} and GenAI-Bench~\citep{jiang2024genai}, both targeting instruction-guided image editing evaluation (pairwise accuracy at $K$=2, 3, 4).

\textbf{Evaluation protocol.}
The 100 preference demonstrations are sampled exclusively from the training split of EditReward-Data, while EditReward-Bench is disjoint and used only for final evaluation.
No additional human feedback is collected, and no model parameters are updated.
Following the EditReward protocol, each predicted $K$-way ranking is converted into all pairwise comparisons.
Therefore, each example with $K=2$, $K=3$, and $K=4$ contributes 1, 3, and 6 comparisons, respectively.
The three EditReward-Bench subsets contain 133, 450, and 888 pairwise comparisons, for 1,471 comparisons in total.

\textbf{Main results.}
Table~\ref{tab:editreward} compares \model against proprietary models (GPT-4o, GPT-5, Gemini, Claude) and open-source baselines (Qwen2.5-VL, MiMo-VL, EditReward) on EditReward-Bench ($K$=2/3/4) and GenAI-Bench. With a frozen Qwen2.5-VL-7B sub-agent, \model achieves 45.7 average accuracy, outperforming all existing baselines, including the strongest open-source reward model EditReward (MiMo) at 44.1 and the strongest proprietary baseline GPT-5 at 42.1. Importantly, this is not simply a backbone advantage: compared under the same Qwen2.5-VL-7B backbone, \model (Qwen) still outperforms EditReward (Qwen) by 3.7 points on average (45.7 vs.\ 42.0). Crucially, this result is obtained without any parameter updates to the underlying VLM and using only 100 preference examples sampled from the EditReward training set for evolution.

The frozen Qwen2.5-VL-7B model scores only 30.3 by itself, so the full system improves it by +15.4 points through the evolved Skills and Tools injected at test time. \model also generalizes well beyond its evolution data: although the Library is evolved only from 100 examples sampled from the EditReward training split, it achieves the best GenAI-Bench accuracy of 67.5, suggesting that the learned Skills and Tools capture general editing-quality criteria rather than benchmark-specific heuristics.

\textbf{Pluggable Sub-Agent.}
The sub-agent is also pluggable. Replacing Qwen2.5-VL-7B with Gemini-2.0-Flash while keeping the same framework yields the best overall average accuracy of 47.4, as well as the best EditReward-Bench performance at $K$=2 and $K$=4. This shows that \model's gains are not tied to a single VLM backbone; instead, the evolved Library transfers across backends and can be combined with stronger sub-agents for further improvement.

\subsection{Performance as Reward Modeling}
\label{subsec:performance_as_reward_modeling}

A reward model is only valuable if it drives genuine improvement in the underlying generative model.
We validate this by using \model as the reward signal in FlowGRPO~\citep{flowgrpo} to fine-tune FLUX.2-klein-base-4B, and evaluating the resulting editor on ImgEdit-Bench~\citep{imgedit} against the base model and an EditReward-trained counterpart under the same FlowGRPO setup.

\textbf{Reward-driven editing improvement.}
As shown in Table~\ref{tab:downstream}, FlowGRPO fine-tuning with \model consistently improves the base model (ImgEdit-Bench Overall: 3.32 $\to$ 3.52), reaching the same overall score as Flux.1 Kontext [dev] despite using a significantly smaller 4B backbone.

\textbf{Comparison under the same FlowGRPO setup.}
Both EditReward and \model are used as reward signals within the same FlowGRPO training pipeline. Under this controlled comparison, \model yields a larger overall improvement on ImgEdit-Bench, raising the base model from 3.32 to 3.52, whereas EditReward reaches 3.45. The two reward signals also lead to different trade-offs across categories: EditReward improves Add and Replace more, while \model delivers stronger gains on Adjust, Extract, Background, and preserves the base-model performance on Compose. Overall, these results indicate that \model provides a more effective training signal than EditReward under the same RL algorithm.

\vspace{-.2cm}
\begin{table*}[t!]
\centering
\renewcommand{\arraystretch}{1.3}
\resizebox{\textwidth}{!}{%
\begin{tabular}{l | w{c}{2.6em} w{c}{2.6em} w{c}{2.6em} w{c}{2.6em} w{c}{2.6em} w{c}{2.6em} w{c}{2.6em} w{c}{2.6em} w{c}{2.6em} w{c}{2.6em}}
\toprule
\multirow{2}{*}{\textbf{Method}} & \multicolumn{10}{c}{\textbf{ImgEdit-Bench}} \\
\cmidrule(lr){2-11}
 & Add & Adjust & Extract & Replace & Remove & Bg. & Style & Compose & Action & Overall \\
\midrule
\rowcolor{proprietary_model_color}
\multicolumn{11}{c}{\textit{Existing Image Editing Models}} \\
\midrule
AnyEdit       & 3.18 & 2.95 & 1.88 & 2.47 & 2.23 & 2.23 & 2.85 & 1.56 & 2.65 & 2.45 \\
UltraEdit     & 3.44 & 2.81 & 2.13 & 2.96 & 1.45 & 2.86 & 3.76 & 1.91 & 2.98 & 2.70 \\
Step1X-Edit  & 3.88 & 3.14 & 1.76 & 3.40 & 2.41 & 3.16 & 4.63 & 2.64 & 2.52 & 3.06 \\
BAGEL            & 3.56 & 3.31 & 1.70 & 3.30 & 2.62 & 3.24 & 4.49 & 2.38 & 4.17 & 3.20 \\
OmniGen2      & 3.57 & 3.06 & 1.77 & 3.74 & 3.20 & 3.57 & 4.81 & 2.52 & 4.68 & 3.44 \\
Ovis-U1          & 4.13 & 3.62 & 2.98 & 4.45 & 4.06 & 4.22 & 4.69 & 3.45 & 4.61 & 4.00 \\
GPT-Image-1  & 4.61 & 4.33 & 2.90 & 4.35 & 3.66 & 4.57 & 4.93 & 3.96 & 4.89 & 4.20 \\
Flux.1 Kontext [dev]                        & 3.76 & 3.45 & 2.15 & 3.98 & 2.94 & 3.78 & 4.38 & 2.96 & 4.26 & 3.52 \\
\midrule
\rowcolor{skills_model_color}
\multicolumn{11}{c}{\textit{Models Tuned with RL}} \\
\midrule
FLUX.2-klein-base-4B                        & 3.27 & 4.00 & 2.04 & 3.07 & 2.90 & 3.47 & 4.26 & 3.19 & 4.44 & 3.32 \\
\quad +RL (EditReward)                      & 4.21 & 4.01 & 2.02 & 3.21 & 2.58 & 3.69 & 4.32 & 2.85 & 4.34 & 3.45 \\
\rowcolor[HTML]{FFF9C4}\quad +RL (\model)                         & 4.03 & 4.27 & 2.24 & 3.12 & 2.19 & 4.22 & 4.44 & 3.19 & 4.22 & \textbf{3.52} \\
\bottomrule
\end{tabular}%
}
\caption{To validate the effectiveness of \model as a reward model, we use it to RL-tune FLUX.2-klein-base-4B and evaluate on downstream image editing benchmarks (ImgEdit-Bench). Compared to EditReward, \model yields more substantial improvements in editing performance.}
\label{tab:downstream}
\end{table*}

\subsection{Analysis}
\label{subsec:analysis}

\begin{figure*}[t]
\centering
\includegraphics[width=0.9\textwidth]{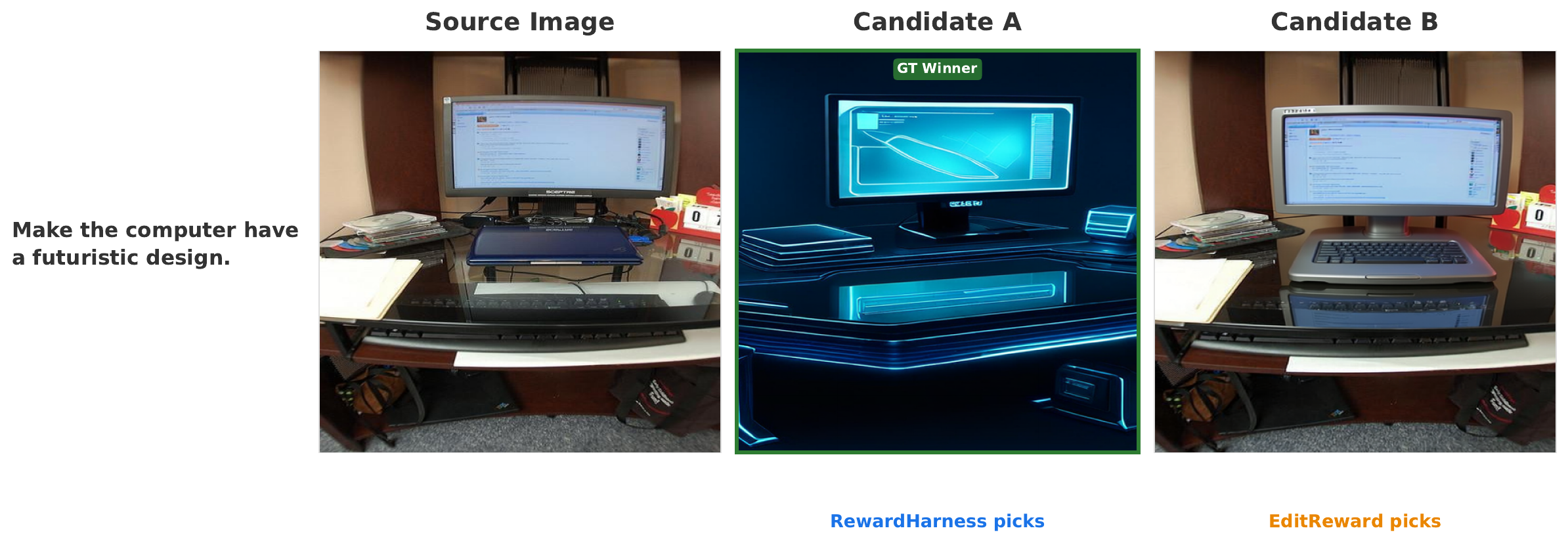}
\caption{\textbf{Reward scoring comparison on EditReward-Bench.} The figure shows a source image, an editing instruction, and two candidate edits (A and B). GT denotes the ground-truth human preference score, \model denotes our predicted score, and ER denotes the EditReward score. \model correctly identifies the human-preferred candidate (marked ``GT Winner''), while EditReward fails.}
\vspace{-.2cm}
\label{fig:scoring}
\end{figure*}

\begin{figure*}[t]
\centering
\includegraphics[width=0.9\textwidth]{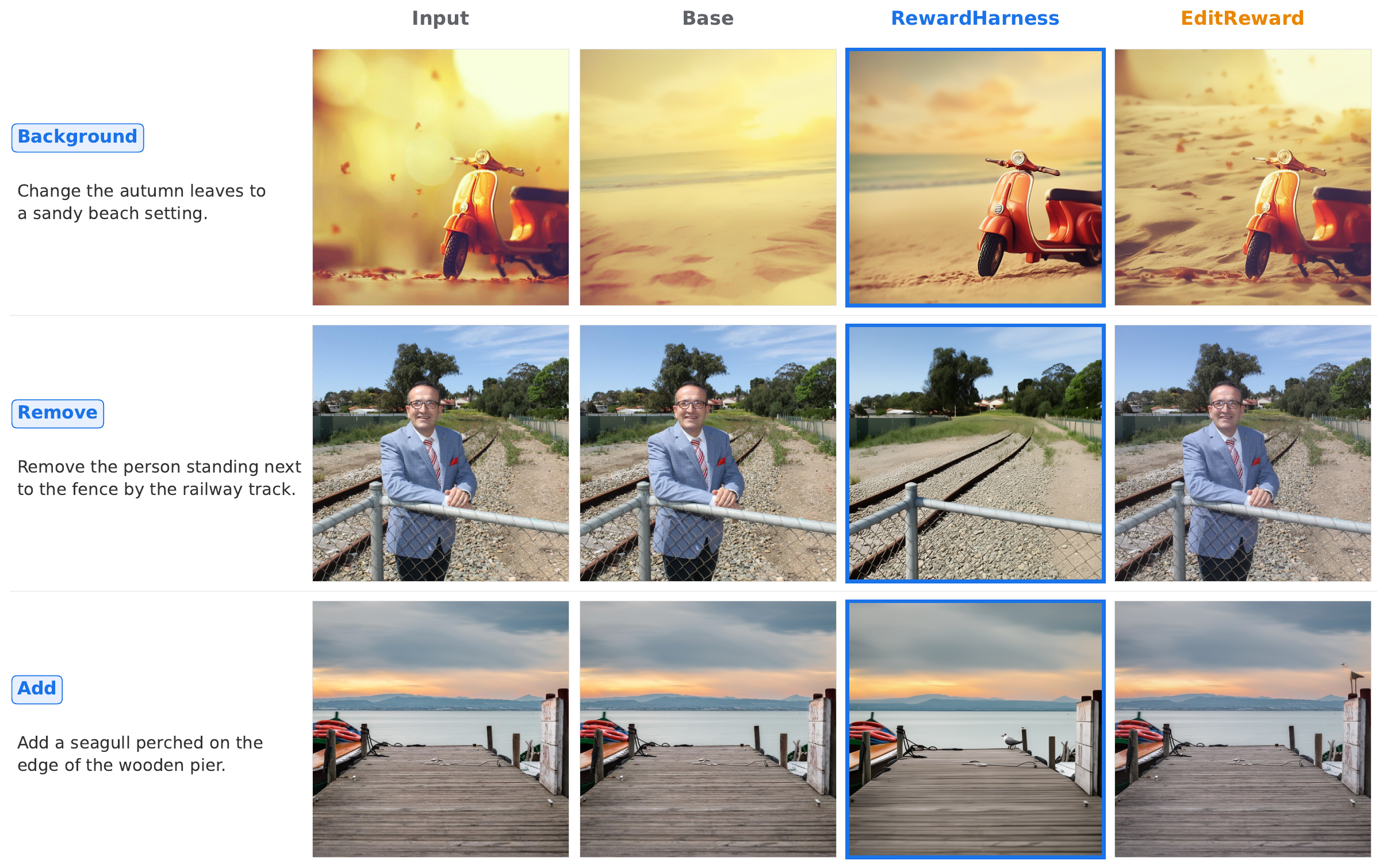}
\caption{\textbf{Qualitative comparison on ImgEdit-Bench.} Each row presents a different editing task with the source image, the base model output (FLUX.2-klein-base-4B), and two RL-fine-tuned variants: \model and EditReward. \model consistently produces edits that faithfully follow the instruction while preserving visual quality and physical consistency, whereas both the base model and the EditReward-trained variant frequently fail to execute the intended edit or introduce artifacts.}
\vspace{-.2cm}
\label{fig:qualitative}
\end{figure*}

\begin{figure*}[t]
\centering
\includegraphics[width=\textwidth]{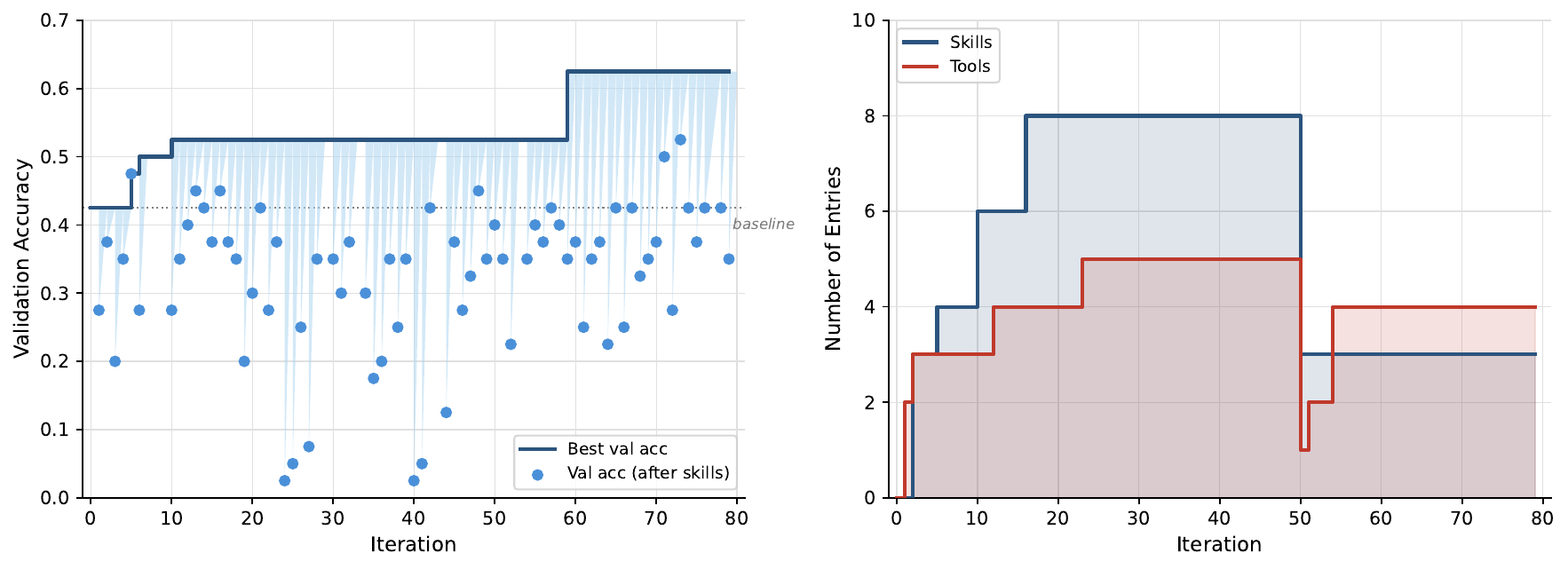}
\caption{\textbf{Self-evolution dynamics over 77 iterations.} \textit{Left:} Per-iteration (dots) and committed (solid line) validation accuracy; the gating mechanism rejects proposals that fail to improve the current committed Library, while the shaded region shows the gap between candidate proposals and the committed state. \textit{Right:} Library size over time. After peaking at 13 entries (8~Skills + 5~Tools), the pruning phase begins around iter~50 and the final committed library reaches 62.5\% accuracy with 7 entries (3~Skills + 4~Tools), a 47\% relative improvement over the 42.5\% baseline.}
\label{fig:evolution}
\end{figure*}

Figure~\ref{fig:evolution} shows the self-evolution dynamics over 77 iterations.
Validation accuracy plateaus at 52.5\% as the library grows to 13 entries (8 Skills + 5 Tools), then improves after the pruning phase begins around iteration~50. The final committed library at iteration~69 reaches 62.5\% validation accuracy with 7 entries (3 Skills + 4 Tools).
Figure~\ref{fig:library_pie} (Appendix~\ref{app:library_case_study}) breaks down library composition at three key stages, illustrating how the system converges to a leaner configuration.
We further examine \model's behavior qualitatively.
Figure~\ref{fig:scoring} shows five reward scoring examples from EditReward-Bench: \model correctly identifies the human-preferred candidate in all cases, while EditReward consistently fails.
Figure~\ref{fig:qualitative} compares RL-tuned editing outputs, showing that the \model-trained variant faithfully executes editing instructions while the base model and EditReward-trained variant frequently fail (see Appendix~\ref{app:qualitative} for additional examples).

\section{Related Work}
\label{sec:related_work}

\textbf{Reward models for visual generation.}
Existing reward models, including ImageReward, PickScore, VisionReward, EditReward, VideoScore2, and ImagenWorld, rely on supervised fine-tuning from tens of thousands of human preference comparisons~\citep{imagereward,kirstain2023pick,xu2024visionreward,editreward,luo2025editscore,he2025videoscore2,sani2026imagenworld}.
\model learns from only $\sim$100 demonstrations by shifting adaptation from parameter updates to test-time context evolution. \textbf{Self-evolving agents.} Context-based self-evolving methods (Reflexion, ExpeL, Voyager, SkillRL, EvolveCoder) keep model weights fixed and evolve prompts, memories, or reusable skills~\citep{shinn2023reflexion,zhao2024expel,wang2023voyager,xia2026skillrl,ruan2026evolvecoder}.
\model specializes this paradigm to multimodal reward modeling: rather than evolving reasoning for a single agent task, we evolve a composable Skills-and-Tools Library that serves as a reusable evaluator context. \textbf{Tool-augmented LLMs.}
Prior work (ReAct, ToolLLM, Gorilla, VerlTool) focuses on learning when to invoke a \emph{fixed} tool set~\citep{yao2022react,qin2023toolllm,patil2023gorilla,jiang2025verltool,zhang2026watchbefore}.
\model inverts this emphasis: the base VLM remains frozen while the Tools themselves are iteratively created and refined to fit the target evaluation domain. See Appendix~\ref{app:related_work_full} for the full literature discussion.

\section{Limitation}

\model relies on a capable Orchestrator for routing, chain analysis, and Library evolution, and its effectiveness may depend on the underlying model's reasoning ability. The framework is modular with respect to the Orchestrator, but a systematic characterization of the capability, API cost, latency, and deployment trade-offs across Orchestrator backbones remains future work.
Our primary evaluation remains focused on instruction-guided image editing. Although the supplementary MMMU result provides preliminary evidence of transfer beyond this domain, effectiveness on text-to-image generation, video editing, and 3D scene manipulation remains to be established.

\section{Conclusion}
\label{sec:conclusion}

We presented \model, a self-evolving agentic framework that reframes reward modeling as context evolution rather than weight optimization. By keeping the underlying VLM frozen and iteratively refining a compact library of Skills and Tools from only 100 labeled samples, \model achieves 47.4\% average accuracy on editing reward benchmarks, surpassing all proprietary and open-source baselines. When used as a reward signal in FlowGRPO, \model improves editing quality while avoiding the reward hacking observed with supervised alternatives. Our results suggest that structured evaluation knowledge, automatically discovered and refined through self-evolution, can be a viable and data-efficient alternative to large-scale preference annotation.

{
\small
\bibliographystyle{colm2026_conference}
\bibliography{references}
}

\appendix

\tcbset{
  casebox/.style={
    colback=gray!6,
    colframe=gray!40,
    colbacktitle={rgb:black,1;white,3},
    coltitle=white,
    fonttitle=\bfseries\small,
    fontupper=\footnotesize\ttfamily,
    boxrule=0.5pt,
    arc=3pt,
    left=6pt, right=6pt, top=4pt, bottom=4pt,
    toptitle=2pt, bottomtitle=2pt,
  }
}

\section{Related Work (Extended)}
\label{app:related_work_full}

\subsection{Reward Models for Visual Generation}
Reward modeling has become a standard approach for aligning visual generators with human preferences. Representative methods include ImageReward, PickScore, VisionReward, UnifiedReward, and OneReward for image generation~\citep{imagereward,kirstain2023pick,xu2024visionreward,unifiedreward,gong2025onereward}, VideoScore for video generation~\citep{he2024videoscore}, and EditReward/EditScore for image editing~\citep{editreward,luo2025editscore}. Despite architectural differences, these methods rely on supervised learning from large-scale human preference data, often ranging from tens of thousands to hundreds of thousands of comparisons~\citep{kirstain2023pick,richhf,editreward}. In contrast, our method learns an image-edit evaluator from only 100 demonstrations by shifting adaptation from parameter updates to test-time context evolution.

\subsection{Self-evolving Agents}
Self-evolving agents improve through feedback from their own interactions, either by updating model parameters (\emph{weight-based}) or by refining textual artifacts (\emph{instruction-/context-based}). Weight-based methods iteratively improve behavior by fine-tuning on self-generated data, as in STaR~\citep{star}, SPIN~\citep{chen2024self}, self-rewarding language models~\citep{yuan2024self}, and related multi-round self-improvement frameworks~\citep{he2025rise,wei2025swe}. In contrast, instruction- and context-based approaches keep model weights fixed and instead evolve prompts, memories, rules, or reusable skills, as in Reflexion~\citep{shinn2023reflexion}, ExpeL~\citep{zhao2024expel}, SCOPE~\citep{pei2025scope}, ACE~\citep{zhang2025agentic}, EvolveR~\citep{wu2025evolver}, Voyager~\citep{wang2023voyager}, SkillRL~\citep{xia2026skillrl}, Agent0~\citep{xia2025agent0}, Agent0-VL~\citep{liu2025agent0}, and SimpleMem~\citep{liu2026simplemem}. Our method belongs to this latter family, but specializes it to multimodal reward modeling: we freeze the underlying VLM and iteratively evolve \emph{Skills} and \emph{Tools} for image-edit evaluation from only 100 labeled samples.

\subsection{Tool-augmented Large Language Models}
Tool-augmented LLMs extend model capabilities by learning when and how to invoke external tools. Representative examples include ReAct~\citep{yao2022react}, Gorilla~\citep{patil2023gorilla}, ToolLLM~\citep{qin2023toolllm}, ReTool~\citep{feng2025retool}, ToolkenGPT~\citep{hao2023toolkengpt}, and broader agent frameworks such as CoALA~\citep{sumers2023cognitive}. In these settings, the tool set is typically assumed to be fixed and learning focuses on the invocation policy. Our method inverts this emphasis: the base models remain frozen, while the \emph{Skills} and \emph{Tools} themselves are iteratively created and refined for the target evaluation task.

\section{Additional Experiments}

\subsection{Data Efficiency Comparison}
\label{app:data_efficiency}

Figure~\ref{fig:paradigm_comparison} contrasts the two paradigms side by side.
The conventional approach (top) requires a large-scale human preference dataset (100k+ annotations) to train a reward model via supervised fine-tuning before any RL alignment can take place. This process is expensive, slow, and infeasible for black-box API models.
\model (bottom) eliminates both the data collection and the fine-tuning stages.
Given only $\sim$100 preference demonstrations as calibration examples, an Orchestrator selects the most relevant skills and tools from a self-maintained library, and a Sub-Agent applies them to produce an interpretable, step-by-step reward score.
Because no gradient updates are made to any VLM, the framework works equally well with closed-source API models and can be deployed immediately on a new domain by providing a small set of labeled examples.

\subsection{Out-of-Domain Transfer on MMMU}
As supplementary evidence outside image editing, we apply the same self-evolution loop to MMMU~\citep{yue2023mmmu}, using Claude Sonnet 4.6 as both the Orchestrator and Sub-Agent.
The Orchestrator diagnoses errors on MMMU-dev and evolves validation-gated Skills and Tools without updating model weights.
On the held-out MMMU validation set, accuracy improves from 79.6 to 80.7 (+1.1).
We treat this result as preliminary evidence that the framework can transfer beyond image editing.

\FloatBarrier
\subsection{Additional Qualitative Examples}
\label{app:qualitative}
\begin{figure}[!htbp]
\centering
\includegraphics[width=0.65\textwidth]{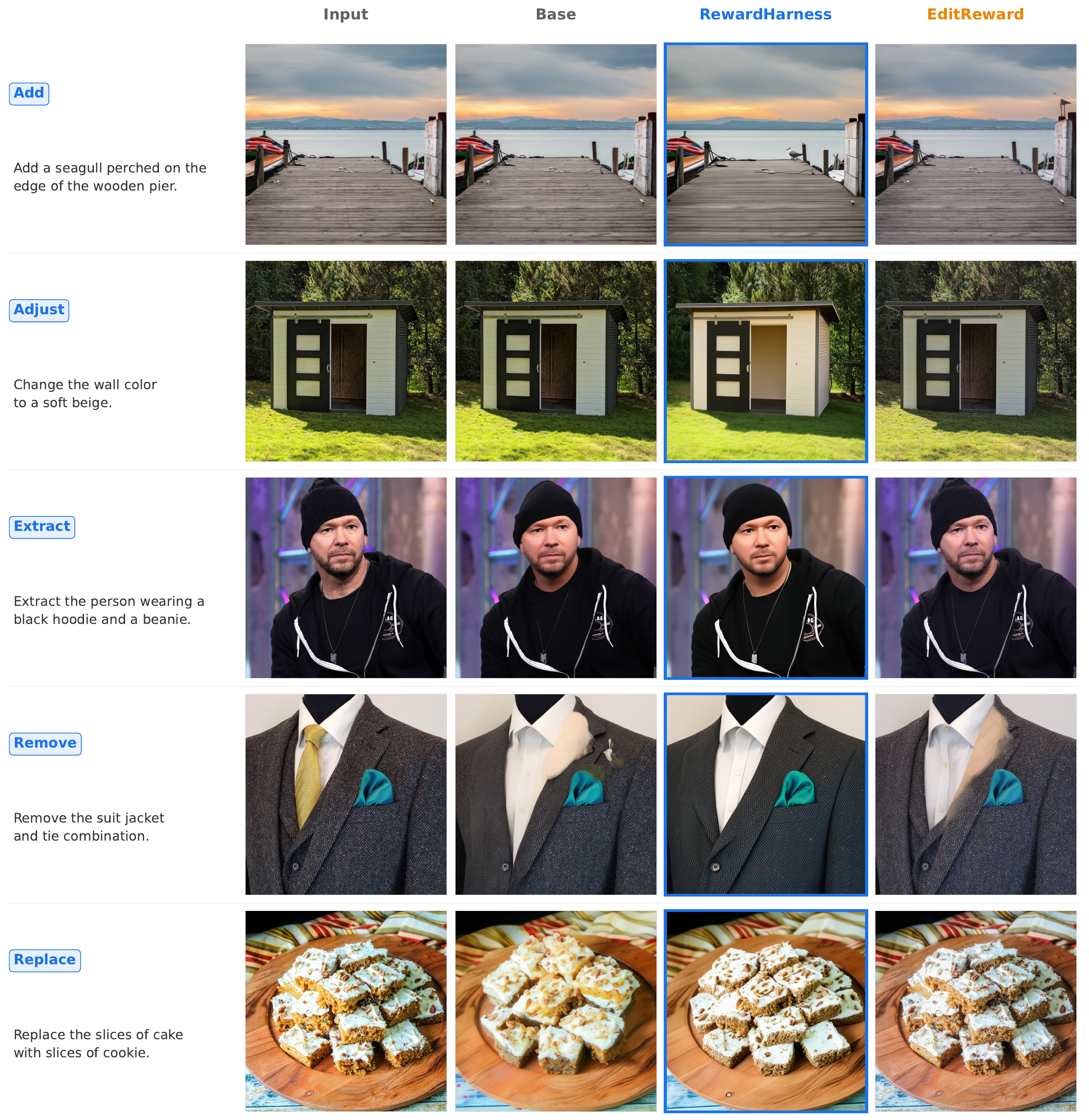}
\caption{\textbf{Additional qualitative comparison on ImgEdit-Bench.} Each row shows a different editing category (Add, Adjust, Extract, Remove, Replace) with the input image, the base model output (FLUX.2-klein-base-4B), and two RL-fine-tuned variants: \model and EditReward.}
\label{fig:qualitative_appendix}
\end{figure}

Figure~\ref{fig:qualitative_appendix} provides additional qualitative examples across five editing categories.

\FloatBarrier

\subsection{Evolution Trajectory}
\label{app:evolution}

Figure~\ref{fig:evolution} shows how validation accuracy and library size co-evolve over 77 iterations of the self-evolution loop on the Gemini-2.0-Flash backbone.
The left panel plots committed validation accuracy (step function) alongside the per-iteration accuracy \emph{after skills are injected} (scatter dots); the triangular fills highlight the temporary performance dip that occurs whenever new skills introduce unseen reasoning patterns. The candidate library is only committed if it improves upon the current committed Library.
Three discrete jumps (iter~5--7, iter~10, iter~58--60) correspond to moments when a newly evolved library state cleared this threshold.
The right panel shows that the library undergoes two structural phases: an \emph{expansion phase} (iters~0--20) where both skills and tools are rapidly added, and a \emph{pruning phase} (iters~50--60) where the Orchestrator replaces the bloated library with a leaner, more reliable configuration that achieves the highest accuracy.

\subsection{Library Case Study}
\label{app:library_case_study}

\begin{figure*}[t!]
\centering
\includegraphics[width=0.85\textwidth]{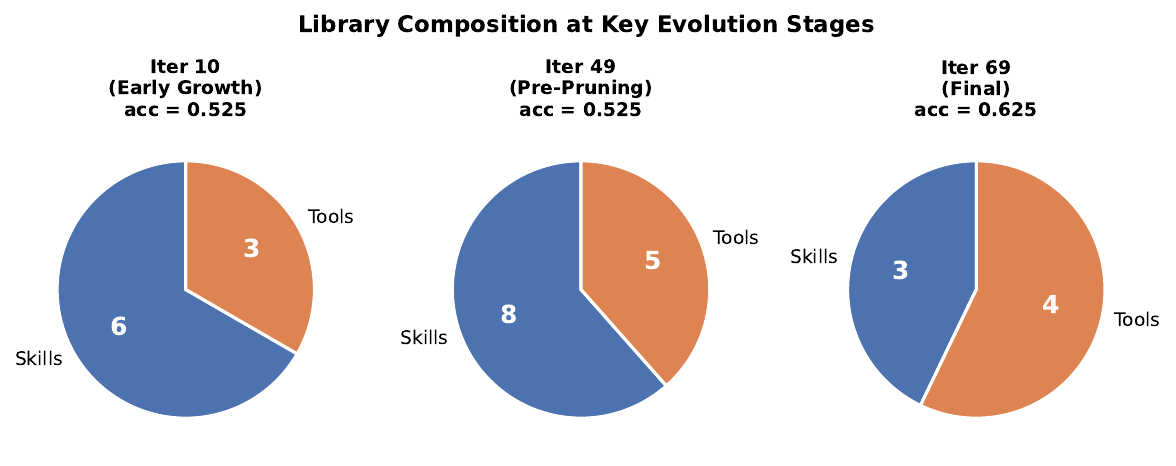}
\caption{\textbf{Library composition at three evolution stages.}
The library grows and then self-prunes: the final configuration (iter~69, val acc~$=0.625$) is leaner than the mid-point peak yet achieves the highest accuracy, with tools outnumbering skills (4 vs.\ 3) as the agent shifts from heuristic guidance to grounded visual verification.}
\label{fig:library_pie}
\end{figure*}

\paragraph{Library composition at key stages.}
Figure~\ref{fig:library_pie} tracks the skills-to-tools ratio at three representative checkpoints.
At iteration~10 the library is skill-heavy (6 skills, 3 tools), reflecting early efforts to encode reasoning heuristics.
By iteration~49 the library peaks at 13 entries (8 skills, 5 tools) before a major pruning event.
The final library at iteration~69 distills to 7 entries (3 skills, 4 tools), with more tools than skills, indicating that the Orchestrator converged on grounding reasoning in explicit visual queries rather than heuristic guidance alone.

\paragraph{Final library contents.}
Table~\ref{tab:library_contents} lists all skills and tools in the final committed library (iteration~69) with their descriptions.
Skills encode \emph{declarative} evaluation heuristics (e.g., describe before judging; allow surrealism if prompted), while Tools provide \emph{procedural} grounding by routing specific sub-queries to a secondary VLM call.

\clearpage
\begin{table}[h]
\centering
\small
\renewcommand{\arraystretch}{1.25}
\begin{tabular}{>{\raggedright\arraybackslash\ttfamily\small\hyphenchar\font=45}p{0.32\textwidth} >{\centering\arraybackslash}p{0.09\textwidth} p{0.50\textwidth}}
\toprule
\normalfont\textbf{Name} & \textbf{Type} & \textbf{Description} \\
\midrule
objective-visual-description-first        & Skill & Mandates describing each image objectively before evaluating, preventing hallucination and position bias. \\
realism-and-artifact-penalties            & Skill & Guides penalizing visual artifacts while explicitly allowing conceptual unrealism when the prompt requests it. \\
style-and-background-transformation-evaluation & Skill & Governs evaluation of background/style changes, enforcing strict foreground preservation and photorealism. \\
\midrule
text-and-ocr-analyzer                     & Tool  & Extracts and verifies text within images to check spelling, placement, and legibility via a secondary VLM call. \\
spatial-and-object-analyzer               & Tool  & Counts objects sequentially and analyzes spatial relationships and orientation via structured JSON output. \\
visual-qa-tool                            & Tool  & Answers targeted visual questions about image content to prevent hallucination and left-right swapping. \\
cultural-and-style-knowledge-oracle       & Tool  & Identifies artistic styles, cultural references, and artist-inspired elements in images. \\
\bottomrule
\end{tabular}
\caption{Final library contents at iteration~69 (3 skills, 4 tools).}
\label{tab:library_contents}
\end{table}

\paragraph{Evolution narrative.}
The analysis summaries logged by the Orchestrator reveal a consistent failure pattern across iterations: the Sub-Agent hallucinates visual content (claiming images are ``completely black'', misreading text, or fabricating object presence), fails on cultural/style knowledge, and over-penalizes conceptually surreal edits that were explicitly requested.
Each evolution step directly targets observed failures. Iteration~1 adds OCR and VQA tools to address text misreading and black-image hallucination; iteration~10 adds a skill to allow surrealism; iteration~60 mandates tool invocation through a new ``tool-usage-mandate'' skill; and the final iteration~69 adds tie-handling guidance for cases where both edited images fail the prompt.
This failure-driven loop mirrors how human annotators iteratively refine rubrics, but operates entirely from model-generated preference comparisons.

\subsection{Skill and Tool Case Studies}
\label{app:skill_case_study}

No zoom-in, crop, or close-up specific skills emerged during evolution.
Spatial and detail-level queries were instead delegated entirely to the \texttt{spatial-and-object-analyzer} Tool, which routes precise counting and layout questions to a structured VLM call.
Below we present three case studies illustrating how the Library evolves toward increasingly targeted guidance.

\begin{figure*}[htbp]
\centering
\includegraphics[width=\textwidth]{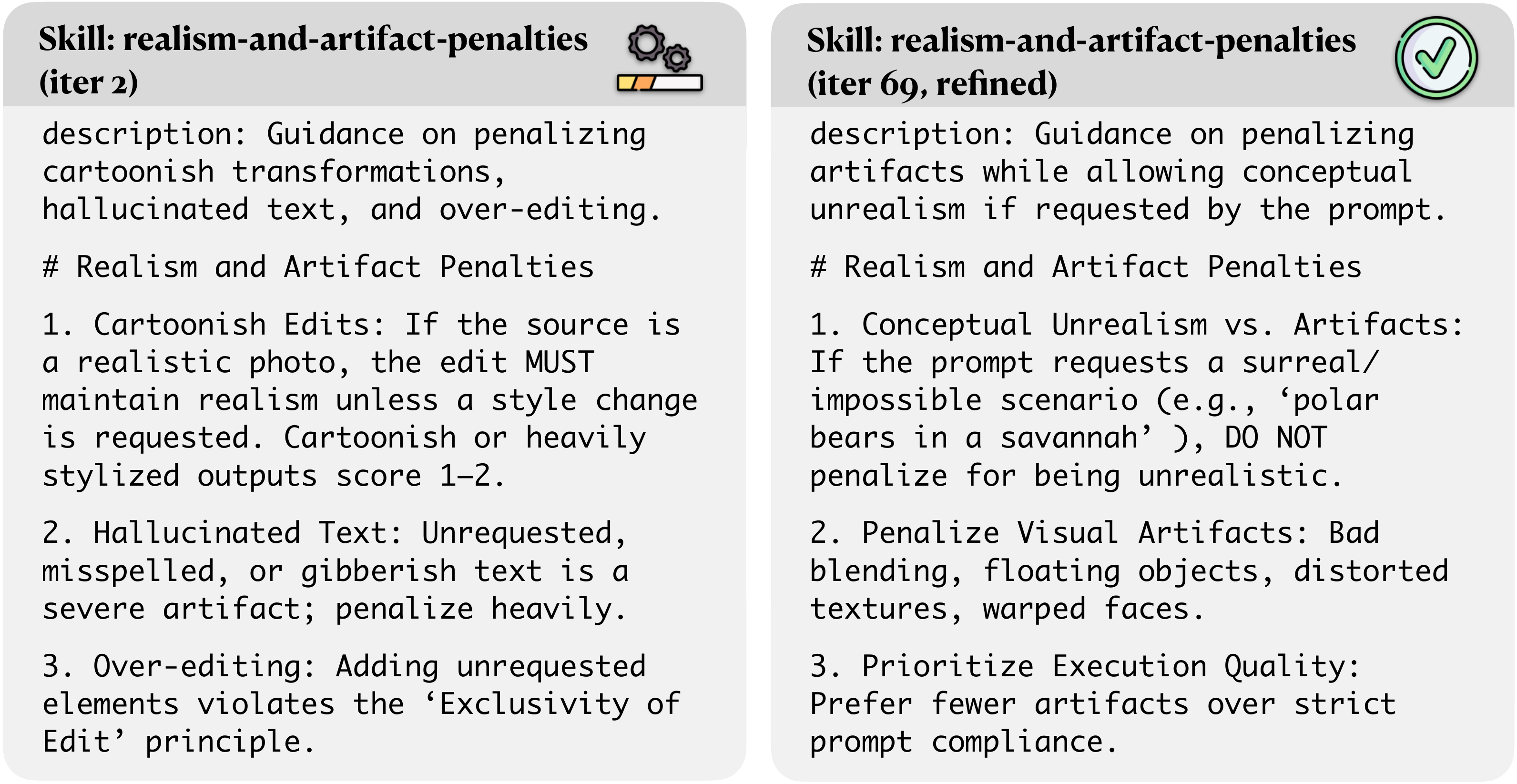}
\caption{Evolution of \texttt{realism-and-artifact-penalties} skill.
Comparison between iteration~2 (left) and iteration~69 (right).
The initial version broadly penalizes cartoonish or unrealistic outputs regardless of intent.
The refined version introduces an explicit carve-out that allows conceptually surreal content when it is requested by the prompt (e.g., ``polar bears in a grassy savannah''), while still penalizing genuine visual artifacts.
This refinement reduces false penalties on prompt-consistent surreal edits and improves alignment between evaluation and user intent.
}
\label{fig:skill_evolution}
\end{figure*}

\medskip
\noindent\textbf{Case 1: Skill Refinement.}\par\noindent
Figure~\ref{fig:skill_evolution} shows \texttt{realism-and-artifact-penalties} at iteration~2 (left) versus iteration~69 (right).
The early version broadly penalizes any cartoonish or unrealistic output; the evolved version adds an explicit carve-out allowing conceptually surreal content \emph{when the prompt itself requests it}, preventing false penalties on prompts such as ``polar bears in a grassy savannah''.

\medskip
\noindent\textbf{Case 2: Tool-Invocation Guidance.}\par
\begin{figure*}[!htbp]
\centering
\includegraphics[width=0.95\textwidth]{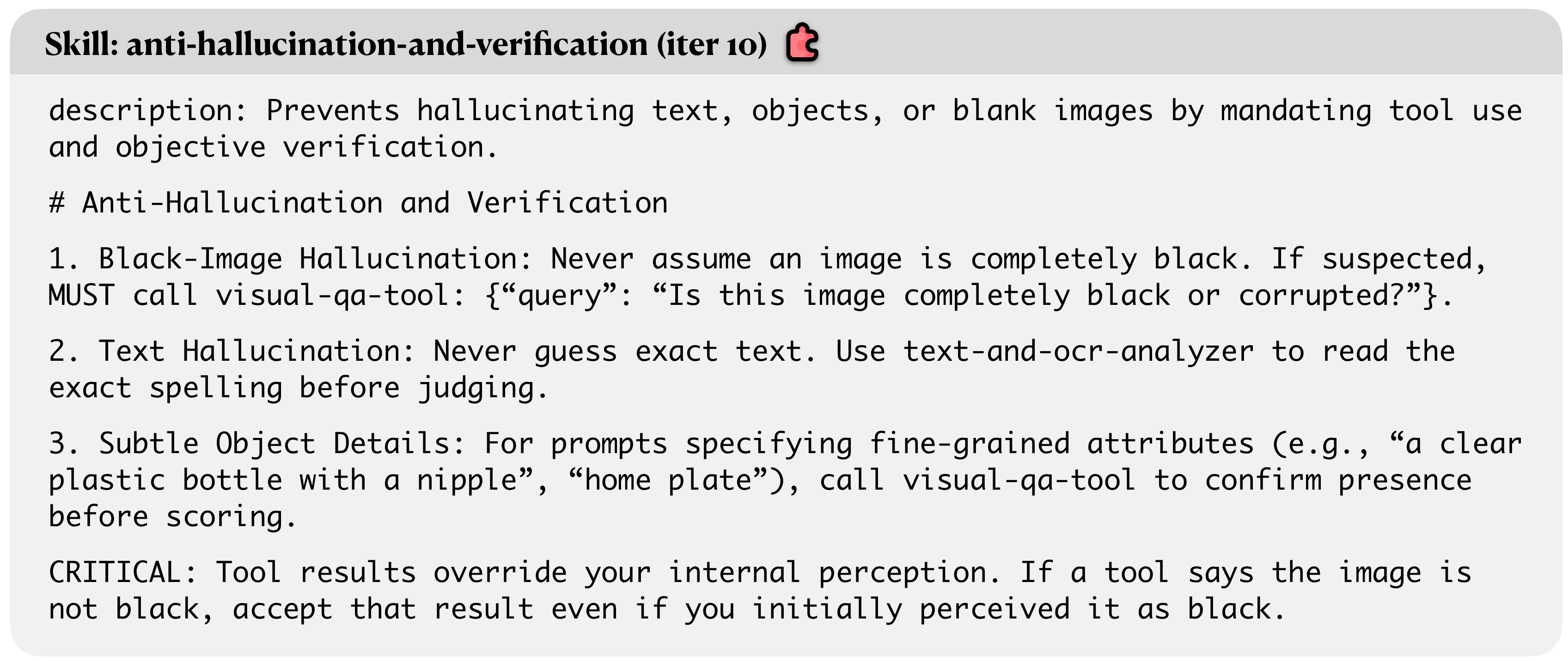}
\caption{The \texttt{anti-hallucination-and-verification} skill (iteration~10) enforces mandatory tool calls for three recurring failure modes. It was later consolidated into tool-level protocols at iteration~60 through the ``tool-usage-mandate'' skill.}
\label{fig:skill_antihallu}
\end{figure*}

\noindent
Figure~\ref{fig:skill_antihallu} shows \texttt{anti-hallucination-and-verification} (iteration~10), a skill that does not evaluate images itself but instead \emph{instructs} the Sub-Agent to route specific queries, including black-image detection, text reading, and object attribute verification, to Tools before forming any judgment.
This pattern represents a Skills$\to$Tools handoff: the Skill specifies \emph{when} to call a Tool; the Tool specification (see below) specifies \emph{how}.

\newpage
\medskip
\noindent\textbf{Case 3: Structured Tool Schema.}\par
\begin{figure*}[!htbp]
\centering
\includegraphics[width=0.95\textwidth]{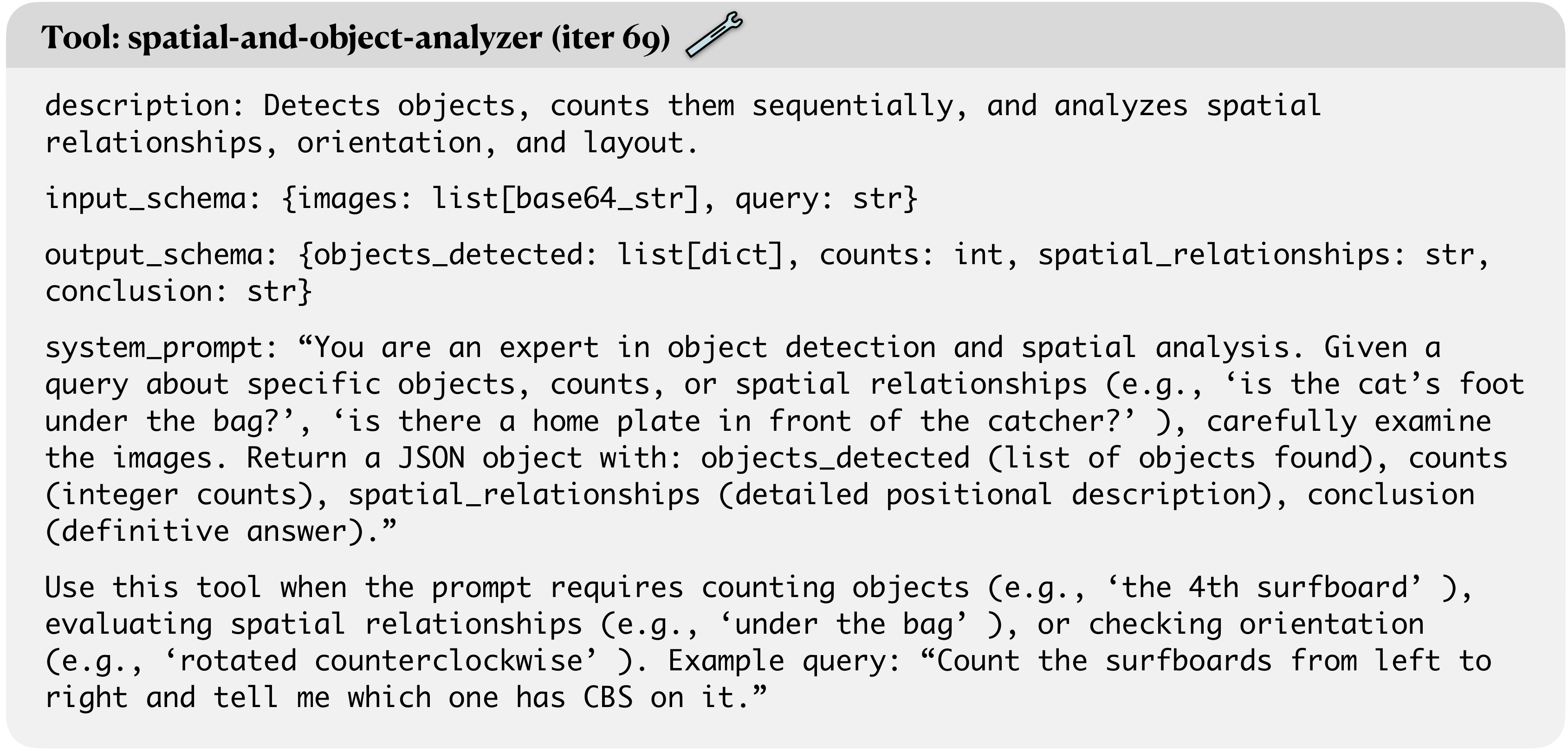}
\caption{The \texttt{spatial-and-object-analyzer} Tool (iteration~69). The typed JSON schema and detailed system prompt ensure the Sub-Agent receives structured, unambiguous grounding for spatial queries.
No dedicated zoom or crop skill was evolved; all fine-grained spatial reasoning was routed through this Tool.}
\label{fig:tool_spatial}
\end{figure*}

\noindent
Figure~\ref{fig:tool_spatial} shows \texttt{spatial-and-object-analyzer}, the tool that handles all spatial, counting, and orientation queries.
Rather than encoding heuristics as text, this Tool issues a secondary VLM call with a typed JSON schema, enabling programmatic grounding for prompts such as ``the 4th surfboard from the left'' and ``is the cat's foot under the bag?'' The primary Sub-Agent consistently hallucinated on such tasks without grounding.

\newpage

\end{document}